\documentclass[conference]{IEEEtran}
\IEEEoverridecommandlockouts

\usepackage{cite}
\usepackage{orcidlink}
\usepackage{tabularx}
\usepackage{pifont}
\usepackage{graphicx}
\usepackage{pstricks}
\usepackage{verbatim}
\usepackage{psfrag}
\usepackage[english]{babel}
 \usepackage{amsmath}
 \usepackage{amssymb}
 \usepackage[font=small]{caption}
  \usepackage{fancybox}
 \usepackage{steinmetz}
\usepackage{subcaption}
\usepackage{bodegraph}
\usepackage{gensymb}
\usepackage{algorithm,algorithmic}
\usepackage{amsfonts}
\usepackage{dsfont}
\usepackage{siunitx}
\usepackage{lscape}
\usepackage{enumerate}
\usepackage{lmodern}
\usepackage[T1]{fontenc}
\usepackage[framed]{matlab-prettifier}
\usepackage{mathtools}
\usepackage{color} 
\usepackage{xcolor,colortbl}
\usepackage{colortbl}
\usepackage{multirow, hhline}
\usepackage{url}
\usepackage{bm}
\usepackage{multicol}
\usepackage{supertabular}
\usepackage{graphicx}
\usepackage{hyperref}
\usepackage[table]{xcolor}

\definecolor{verdemio}{rgb}{0.01, 0.75, 0.24}

\graphicspath{{Images/}}

 \makeatletter
 \newcommand\notsotiny{\@setfontsize\notsotiny\@vipt\@viipt}
 \makeatother

\definecolor{Gray}{gray}{0.9}

\definecolor{verdedd}{rgb}{0.0, 0.5, 0.0}

\definecolor{myorangenew}{rgb}{1 0.49 0}
\definecolor{mygreennew}{rgb}{0.31 0.78 0.47}
\definecolor{mygreynew}{rgb}{0.6 0.6 0.6}
\definecolor{bluD}{rgb}{0.09, 0.45, 0.81}
\definecolor{bluD_l}{rgb}{0.29, 0.65, 1}
\definecolor{nero}{rgb}{0, 0, 0}
\definecolor{bianco}{rgb}{1, 1, 1}

\definecolor{bluG}{cmyk}{100 85 0 0}
\definecolor{verdeG}{cmyk}{75 0 100 0}

\def\q{{\boldsymbol q}}

\DeclareMathSymbol{\boldmu}{\mathord}{letters}{15}

\begin{document}

\title{
Integrated Identification of Collaborative Robots for Robot Assisted 3D Printing Processes

\thanks{The authors are with the Department of Engineering ``Enzo Ferrari'', University of Modena and Reggio Emilia, via Pietro Vivarelli 10, 41125 Modena, Italy, e-mail: \{alessandro.dimauro, davide.tebaldi, fabio.pini, luigi.biagiotti, francesco.leali\}@unimore.it}
\thanks{The work was funded under the National Recovery and Resilience Plan (NRRP), Mission 04 Component 2 Investment 1.5 – NextGenerationEU, Call for tender n. 3277 dated 30/12/2021 Award Number:  0001052 dated 23/06/2022, and was supported by the Italian National Recovery and Resilience Plan (PNRR), Mission 4 “Education and Research”, Component C2, Investment 1.1 “PRIN – Projects of Relevant National Interest”, Project I-SHARM: Intelligent SHared Autonomy for Robotic Manipulation Systems, Project ID 2022NTZRFM, CUP E53C24002600006.}
}

\author{
\IEEEauthorblockN{Alessandro Dimauro }
\IEEEauthorblockA{\textit{Dept. of Eng. “E. Ferrari” (DIEF)}\\
\textit{Unimore, Italy} \\
alessandro.dimauro@unimore.it}

\and
\IEEEauthorblockN{Davide Tebaldi }
\IEEEauthorblockA{\textit{Dept. of Eng. “E. Ferrari” (DIEF)}\\
\textit{Unimore, Italy} \\
davide.tebaldi@unimore.it}

\and
\IEEEauthorblockN{Fabio Pini }
\IEEEauthorblockA{\textit{Dept. of Eng. “E. Ferrari” (DIEF)}\\
\textit{Unimore, Italy} \\
fabio.pini@unimore.it}

\and
\hspace{35mm}
\IEEEauthorblockN{Luigi Biagiotti }
\IEEEauthorblockA{\hspace{35mm}\textit{Dept. of Eng. “E. Ferrari” (DIEF)}\\
\textit{\hspace{35mm}Unimore, Italy} \\
\hspace{35mm}luigi.biagiotti@unimore.it}

\and
\IEEEauthorblockN{Francesco Leali }
\IEEEauthorblockA{\textit{Dept. of Eng. “E. Ferrari” (DIEF)}\\
\textit{Unimore, Italy} \\
francesco.leali@unimore.it}
}

\maketitle

\begin{abstract}
In recent years, the integration of additive manufacturing (AM) and industrial robotics has opened new perspectives for the production of complex components, particularly in the automotive sector. Robot-assisted additive manufacturing processes overcome the dimensional and kinematic limitations of traditional Cartesian systems, enabling non-planar deposition and greater geometric flexibility. However, the increasing dynamic complexity of robotic manipulators introduces challenges related to precision, control, and error prediction. This work proposes a model-based
approach equipped with an integrated identification procedure of the system's parameters, including the robot, the actuators and the controllers.  

We show that the integrated modeling procedure allows to obtain a reliable dynamic model even in the presence of sensory and programming limitations typical of collaborative robots. 
  The manipulator's dynamic model is identified through an integrated five-step methodology: starting with geometric and inertial analysis, followed by friction and controller parameters identification, all the way to the remaining parameters identification. The proposed procedure intrinsically ensures the physical consistency of the identified parameters.
  The identification approach is validated on a real-world case study involving a 6-Degrees-Of-Freedom (DoFs) collaborative robot used in a thermoplastic extrusion process. The very good matching between the experimental results given by actual robot and those given by the identified model shows the potential enhancement of precision, control, and error prediction in Robot Assisted 3D Printing Processes.
  \end{abstract}

\section{Introduction}

Additive Manufacturing (AM) processes have brought a radical change in the manufacturing technologies of parts used in industrial production. The design freedom given by the possibility of building components layer by layer starting from a Computer-Aided Design (CAD) file, in fact, allows for a greater variety of possible geometries, a reduction in material waste and, sometimes, an acceleration of the production cycle \cite{Xue2025,Shahrubudin2019}. 
In particular, in the automotive sector, 3D printing not only allows for the creation of light and complex structures in a short time \cite{Sreehitha2017}, but also offers the possibility of printing parts starting from the most varied 3D geometries and obtaining a systematic archiving of the geometries of components that are no longer available on the market \cite{Wawryniuk2024}. 
However, the intrinsic limit of AM processes lies in the difficulty in adapting them for large-scale production. In fact, the amount of output required by mass applications is today much higher than the range in which AM is economically advantageous. Although this aspect is being overcome thanks to emerging technologies, for example binder jetting processes, AM is still mainly placed in the racing or luxury segments \cite{AlMakky2016}. Furthermore, post-processing requirements \cite{Kumar2025}, anisotropic properties of the printed part due to the particular printing strategy chosen \cite{Adejumo} and limited high-temperature applications play an important role in the choice not to employ AM processes. \\
To overcome these limitations, recent research has focused on robot-assisted additive manufacturing in which an industrial or collaborative manipulator replaces the traditional three-axis Cartesian system. This step allows for extended workspace, non-planar deposition and the integration of advanced sensors for in-line quality control \cite{Pires2022,Zhang2023}. 
However, this evolution introduces new critical issues: articulated arms present a lower stiffness than Cartesian systems and a strongly coupled dynamics between joints, with significant effects on positional accuracy during high-speed trajectories \cite{DeMarzi2023}. \\
Recent literature highlights a growing attention towards the use of digital twins and advanced dynamic models to predict and compensate errors induced by manipulator dynamics \cite{Xiang2024,Zhang2020}. 
An interesting aspect is the use of model-based approaches, which allow modeling complex multibody systems coupled with electrical actuation and control systems as systems of differential-algebraic equations.  
The use of model-based approaches allows for easy symbolic analysis, facilitating parameter traceability \cite{Bardaro2017}. Such model-based approaches have to be used in conjunction with effective parametric identification methods \cite{Khanesar2023,Sung2025}.  

Parameters identification approaches used for industrial robots include regression techniques \cite{Hollerbach2008ModelIdentification,Janot2014GenericInstrumentalVariable}, exploiting the fact that the robot dynamics is linear w.r.t. a set $\boldsymbol{\pi} \in \mathcal{R}^p$ of dynamic parameters (masses and inertias):
\begin{equation}\label{ident_prob}    
\boldsymbol{\tau}=\boldsymbol{\Phi}(\boldsymbol{\q},\boldsymbol{\dot{\q}},\boldsymbol{\ddot{\q}})\boldsymbol{\pi},
\end{equation}
where $\boldsymbol{\tau}\in \mathcal{R}^n$ is the torque vector and $\boldsymbol{\Phi}(\boldsymbol{\q},\boldsymbol{\dot{\q}},\boldsymbol{\ddot{\q}})\in \mathcal{R}^{(n\times p)}$ is the dynamic regressor. However, these methods do not consider the set of feasible numerical values for the
dynamic parameters $\boldsymbol{\pi}$, which is relevant to ensure the physical consistency of the model \cite{Mata2005DynamicParameterIdentification}. This aspect has been addressed in the literature in different ways, e.g. solving the identification problem using semidefinite
programming (SDP) techniques \cite{Sousa2014PhysicalFeasibility} while further advancements include the addition of the triangle
inequality of the tensors of inertia \cite{Sousa2019InertiaTensorLMI,Traversaro2016PhysicalConsistency}. The approach proposed in this paper does not rely on the imposition of constraints to the identification of dynamic coefficients \cite{Gaz2019FrankaIdentification}, nor does it consist in identifying all the parameters at once as typically done by other approaches. Instead, we address the identification through a five-step methodology: starting with geometric and inertial analysis, followed by friction and controller parameters identification, all the way to the remaining parameters identification. This approach has the advantage of decoupling the complex identification problem \eqref{ident_prob} into a sequence of subproblems, and exploits in the initial phase the geometric/inertial information that is typically already available from, e.g., the CAD file. In doing so, the physical consistency of the identified parameters is intrinsically guaranteed. Furthermore, the approach we propose does not only include the identification of the manipulator dynamics, as typically done by other approaches, but also the identification of the actuators dynamics and of the controllers.

The identification approach is validated on a real-world case study involving a 6-DoFs collaborative robot used in a thermoplastic extrusion process. The close agreement between the actual robot experiments and the outcomes generated by the identified model highlights the capability to enhance accuracy, process control, and error prediction in Robot-Assisted 3D Printing.

\section{Methods and Tools}

The first step consists in the development of the dynamic model of the robot manipulator, including the actuators dynamics, the transmission dynamics and the dynamics of the controllers. Such aspect is addressed in Sec.~\ref{model}. Subsequently, the proposed integrated identification procedure is presented in Sec.~\ref{identif_sect}. Finally, the results of the identification and the model validation through experiments are addressed in Sec.~\ref{param_id_resultd} and in Sec.~\ref{Model_Verf}, respectively.

\section{Dynamic Modeling of Robotic Manipulators, Actuators and Controllers}\label{model}

\begin{figure}[t]
 \centering\includegraphics[clip,width=0.959\linewidth]{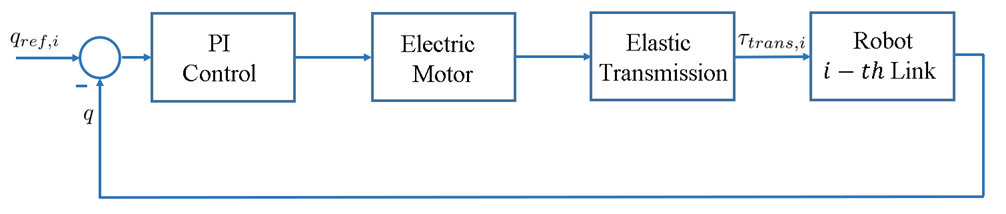}
    \caption{Structure of the feedback system composed of the controller, the electric motor, the elastic transmission and the $i$-th link of the robot manipulator.}
\label{fb_syst}
\vspace{-4.2mm} 
 \end{figure}

The 
multibody dynamic model of the manipulator under analysis consists of a chain of $n$ rigid bodies connected by ideal revolute joints, each actuated by an electric motor, modeled as the electric dynamics represented by stator inductance and resistance, by the torque constant performing electrical to mechanical energy conversion, and by the rotor mechanical dynamics.
The connection between the motor and the associated joint is realized through a spring-damper transmission. Each motor receives as input a voltage generated by a linear SISO (Single Input Single Output) controller, which receives as feedback the angular position of the corresponding joint. The end-effector is modeled as a rigid body fixed to the last link of the manipulator. The structure of the feedback system associated with each link is shown in Fig. \ref{fb_syst}.

Concerning the $i$-th link, let $q_{m,i}$ denote the rotor angular position, $\dot{q}_{m,i}$ the rotor angular velocity, $\ddot{q}_{m,i}$ the rotor angular acceleration, 
$\tau_{m,i}$ the motor-generated torque, $\tau_{f,i}$ the friction torque, $\tau_{trans,i}$ the elastic transmission torque, and $\tau_{ext,i}$ the external torque. By denoting as $M(q)$ the mass matrix, which is symmetric and positive definite, $C(\dot{q},q)$ as the Coriolis and centrifugal matrix, and $g(q)$ as the gravitational force vector, the Euler–Lagrange equations for the manipulator can be written as:

\begin{equation}
\sum_j \!\! \left[ M_{ij}(q)\ddot{q}_j \!+ \!C_{ij}(q,\dot{q})\dot{q}_j \right]
+ g_i(q) + \tau_{f,i} + \tau_{ext,i} - \tau_{trans,i} = 0,
\label{first}
\end{equation}
where index $i$ denotes the $i$-th link, and $j=1,\,\ldots,\,n$, with $n$ the number of DoFs of the system. The friction term $\tau_{f,i}$ can be modeled according to the following Stribeck plus viscous formulation:

{\footnotesize
\begin{equation}
\tau_{f,i} \!=\!\!
\begin{cases}
\!\!\left[\!
F_{c,i} \!\!+\! (F_{s,i} \!\!-\! F_{c,i})\!\!
\left(\!\!
\dfrac{v'_{s,i} \!- \!|\dot{q}_i|}{v'_{s,i}}
\!\right)\!\!
\right]\!\!
\operatorname{sign}(\dot{q}_i)
\!\!+\!\! F_{v,i}\dot{q}_i
\!\!&\! \text{if } |\dot{q}_i|\! \le\! |v'_{s,i}| \\[10pt]

F_{c,i}\operatorname{sign}(\dot{q}_i)
+ F_{v,i}\dot{q}_i
\!\!&\! \text{if } |\dot{q}_i| \!>\! |v'_{s,i}|
\end{cases},
\label{sec}
\end{equation}}
where $F_{c,i}$ is the Coulomb friction torque, 
$F_{v,i}$ is the viscous friction coefficient, 
$F_{s,i}$ is the static friction torque, 
and $v'_{s,i}$ is the Stribeck velocity defining the transition between static and Coulomb friction.

 Concerning the electric motor, let $I_i$ denote the motor armature current, $V_i$ the motor armature voltage, $K_{t,i}$ the motor torque constant, $K_{b,i}$ the motor back-emf constant, $R_i$ the motor armature resistance, $L_i$ the motor armature inductance, and $J_{m,i}$ the rotor inertia. The electrical and mechanical dynamics of the electric motor can be written as:

\begin{equation}
V_i - R_i I_i -K_{b,i}\dot{q}_{m,i}=L_i \frac{dI_i}{dt}
\label{thi}
\end{equation}

\begin{equation}
J_{m,i}\ddot{q}_{m,i} = \tau_{m,i} - \frac{\tau_{trans,i}}{r_i}, \hspace{4mm} \mbox{where}\hspace{4mm} \tau_{m,i} = K_{t,i} I_i.
\label{for}
\end{equation}

Concerning the transmission system, let $K_{s,i}$ denote the torsional stiffness of the transmission, $K_{c,i}$ the transmission damping coefficient, and $r_i$ the transmission ratio. The elastic transmission dynamics can be written as:

\begin{equation}
\tau_{trans,i} = K_{s,i}\left(\frac{q_{m,i}}{r_i} - q_i \right)
+ K_{c,i}\left(\frac{\dot{q}_{m,i}}{r_i} - \dot{q}_i \right).
\label{five}
\end{equation}

Finally, the transfer function of the PI controller is the following:

\begin{equation}
\dfrac{\tau_{m,i}(s)}{\tilde{q}_i(s)} =  \frac{K_{p,i}(1 + T_{z,i}\, s)}{s},
\label{seven}
\end{equation}
where $s$ is the Laplace variable, $\tau_{m,i}(s)=\mathcal{L}(\tau_{m,i})$ and $\tilde{q}_i(s)=\mathcal{L}(\tilde{q}_i)$ are the Laplace transform of the motor torque $\tau_{m,i}$ and of the tracking error $\tilde{q}_i=q_{ref,i} - q_i$, respectively.

\section{Integrated Parameter Identification Procedure}\label{identif_sect} 

\begin{figure*}[t]
 \centering\includegraphics[clip,width=0.959\linewidth]{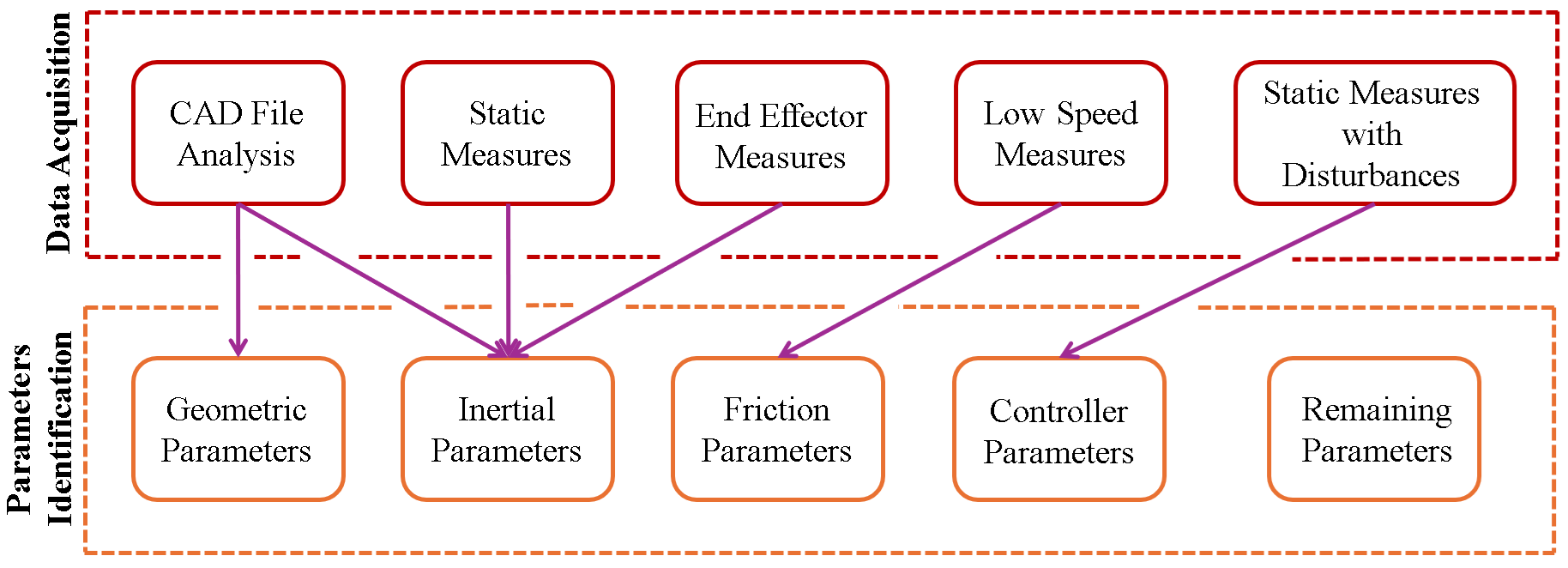}
    \caption{Workflow of the integrated parameters identification procedure. 
    }
\label{figure_2}
\vspace{-4.2mm} 
 \end{figure*}

The workflow of the proposed integrated parameters identification procedure is outlined in Fig. \ref{figure_2}.

Therefore, the identification procedure consists in dividing the parameters to be identified into the different subsets of Fig. \ref{figure_2}: geometric, inertial, friction, controller and remaining parameters.

\subsection{Step 1: From CAD file to Geometric and first-attempt Inertial Parameters.}\label{step1}

The identification of geometric and inertial parameters starts from the awareness that 3D models of the various parts of the robot are normally available, and such knowledge should therefore be exploited. 
Through the geometric analysis tools available in any CAD software, it is possible to calculate the total volume of the assembly and of the individual parts. Therefore, using the total mass of the robot, often publicly available or experimentally measurable, it is possible to derive the average density of the assembly and apply it to the individual parts.

At this stage, the geometric parameters of the links (distance vector between the two joints) and of the joints (unit vector of the rotation axis), as well as first-attempt inertial parameters of the moving links (center of mass position vector and inertia tensor), are identified. In the case of a 6-DoFs robot, at the end of this phase a total of 39 geometric parameters (3 per joint and 3 per link) and 60 inertial parameters (10 per moving link) are obtained.

 \subsection{Step 2: From Static and End Effector Measures to Identification Refinement of Inertial Parameters.}\label{step2}

 To obtain more precise information on some inertial parameters (masses, center of mass positions), static measurements can be performed in strategically calculated poses. A way to obtain faster measurements with fewer poses consists in identifying only the most relevant inertial parameters, which have a direct and appreciable effect on the static joint torque. Therefore, poses are selected so that the parameters to be inserted into the model can be directly derived from the measured torques using explicit formulas. This eliminates the need for an iterative procedure, which may be very time-consuming.

For completeness of the model, it is necessary to include the geometric and inertial data related to the end-effector. It should be noted that static friction may contaminate the initial phase of each pose measurement; however, if 
this phenomenon is observed, it is useful to eliminate the contribution of $\tau_{f,i}$ in \eqref{sec} by keeping the robot in slow and small oscillations around the prescribed pose.

Eliminating $\tau_{f,i}$ and following the information obtained from appropriate measurements of $\tau_{m,i}$ in \eqref{for} in static poses, the dynamic equilibrium equation of the manipulator \eqref{first}, for $\dot{q}_i = 0$ and $\ddot{q}_i = 0$, becomes:
\[g_i(q) + \tau_{ext,i} - \tau_{trans,i} = 0.
\]
Assuming the robot operates without external disturbance, i.e. $\tau_{ext,i}=0$, one can write:
\[g_i(q) = \tau_{trans,i}.
\]
Similarly, the equilibrium condition of the motor mechanical dynamics \eqref{for} becomes:
\begin{equation}
\tau_{m,i} = \frac{\tau_{trans,i}}{r_i}.
 \label{static_eq_motor}
 \end{equation}
Let $m_k^{CAD}$ and $I_k^{CAD}$ denote the $k$-th link mass and the $k$-th link moment of inertia available from CAD, that are the first-attempt values assigned at {\it Step 1} in Sec. \ref{step1}. These values are updated as follows:
\begin{equation}
m_k=\lambda_k m_k^{CAD} \hspace{4mm} \mbox{and} \hspace{4mm}
I_k=\lambda_k I_k^{CAD}, 
 \label{lambda_k_eq}
 \end{equation}
where $\lambda_k$ is a correction coefficient calculated from static measurements.

 \subsection{ Step 3: From Low Speed Measures to Identification of Friction Parameters.}\label{step3}

The identification of the friction parameters is performed without a force/torque sensor at the robot base. The method exploits symmetric periodic trajectories with sufficiently long constant-velocity segments in both motion directions, i.e. with positive and negative velocities, so that friction can be isolated by subtraction while inertial, Coriolis/centrifugal, and gravitational effects are eliminated, as described in the following.

Let us start from the manipulator dynamic model in Eq.~\eqref{first} and consider the motion of one joint $i$ at a time, while the others remain at rest. In the constant-velocity segments, 
$\ddot q_i = 0$ holds
and, since the other joints are not moving, the inertial contribution $\sum_j M_{ij}(q)\ddot q_j$ vanishes.

Due to the symmetry of the prescribed motion, the two constant-velocity segments occur at the same configuration but with opposite velocity. Therefore:
(i) the gravitational term $g_i(q)$, depending only on the configuration, is identical in both segments;
(ii) the Coriolis/centrifugal term $\sum_j C_{ij}(q,\dot q)\dot q_j$ is also identical in both segments;
(iii) the external torque $\tau_{ext,i}$ is considered to be the same.

Observation (ii) arises from the fact that, although $C_{ij}(q,\dot q)$ is linear in $\dot q$, the complete Coriolis/centrifugal contribution $\sum_j C_{ij}(q,\dot q)\dot q_j$ is quadratic in velocity and therefore even:
\[
C(q,-\dot q)(-\dot q) = C(q,\dot q)\dot q.
\]
Consequently, it cancels when subtracting the two constant-velocity equations.

Let $r \in \{1,\,2\}$ denote the two constant-velocity segments. Under the above assumptions, the dynamic equation of the manipulator reduces to
\begin{equation}
\label{robot_consid}
\left( \sum_j C_{ij}(q,\dot q)\dot q_j \right)^{(r)}
+ g_i(q)
+ \tau_{f,i}^{(r)}
- \tau_{trans,i}^{(r)} = 0.
\end{equation}

Subtracting the second equation (Eq. \eqref{robot_consid} with $r=2$) from the first (Eq. \eqref{robot_consid} with $r=1$) yields
\begin{equation}
\tau_{f,i}^{(1)} - \tau_{f,i}^{(2)}
=
\tau_{trans,i}^{(1)} - \tau_{trans,i}^{(2)}.
\end{equation}

Since the friction torque is an odd function of velocity,
\[
\tau_{f,i}(-\dot q_i) = -\tau_{f,i}(\dot q_i),
\]
it follows that
\begin{equation}
2|\tau_{f,i}| =
|\tau_{trans,i}^{(1)} - \tau_{trans,i}^{(2)}|.
\end{equation}

Therefore, using the relation between $\tau_{trans,i}$ and the motor torque (Eq.~\eqref{static_eq_motor}), the friction characteristic $\tau_{f,i}(\dot q_i)$ can be reconstructed from the experimental data, and the parameters of the friction model (Eq.~\eqref{sec}) can be identified.

 \subsection{Step 4: From Static Measures with Disturbances to Identification of Controller Parameters.}\label{step4}

Static measurements under disturbance play a fundamental role in identifying controller-related parameters.

The measurements are performed at a constant set point $q_{\mathrm{ref},i}$ for each joint (robot at rest), while an external torque disturbance $\tau_{\mathrm{ext},i}$ is artificially applied to each joint, with a magnitude that is varying in time but not large enough to trigger the robot’s safety system (which would immediately stop all operations, including the recording of sensor outputs).

The identification of the parameters $K_{p,i}$ and $T_{z,i}$ of the controller in Eq.~\eqref{seven} for each joint is subsequently carried out through an iterative search process using MATLAB’s System Identification Toolbox, which allows to determine the optimal values for the controller parameters.
\begin{figure}[t]
 \centering\includegraphics[clip,width=0.859\linewidth]{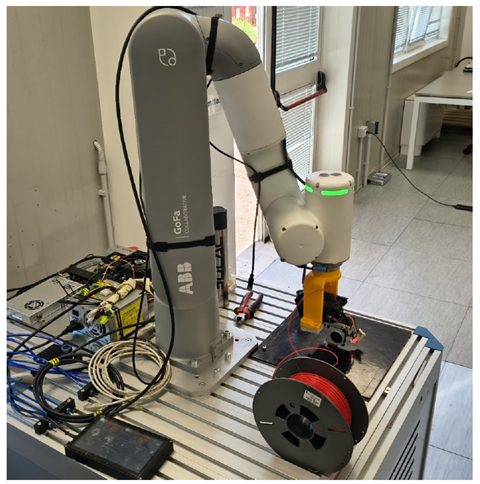}
    \caption{Experimental setup involving  an ABB GoFa CRB15000
collaborative robot for 3D printing process.}
\label{exp_setup}
\vspace{-4.2mm} 
 \end{figure}

The need to proceed in this way arises from a limitation common to all collaborative robots, namely the impossibility of prescribing and executing abrupt or excessively fast movements, for safety and operator protection reasons. It is therefore impossible to perform a direct frequency-domain analysis that provides meaningful information, since oscillations that are too fast would be blocked before reaching the control system. The proposed solution allows these limits to be overcome and a relationship between position error and control action to be properly identified. 
\begin{table*}[h!]
\centering
\begin{tabular}{c|cccccccc}
\hline
\textbf{$k$-th Link} & \textbf{0} & \textbf{1} & \textbf{2} & \textbf{3} & \textbf{4} & \textbf{5} & \textbf{6} & \textbf{End Effector} \\
\hline
$x_{g,k}$ (m) & 0 & -3.794E-06 & 1.084E-05 & 1.282E-02 & 1.605E-01 & -1.470E-02 & 1.315E-02 & 7.300E-02 \\
$y_{g,k}$ (m) & 0 & 1.030E-02 & -4.993E-02 & 9.437E-02 & 4.197E-02 & -7.954E-02 & 2.453E-04 & 0 \\
$z_{g,k}$ (m) & 8.610E-02 & 7.048E-02 & 2.220E-01 & 2.470E-02 & 0 & 4.276E-02 & -3.500E-04 & 5.500E-03 \\
\hline
$x_{k}$ (m) & 0 & -7.366E-07 & 7.105E-07 & 9.600E-02 & 3.740E-02 & 6.800E-02 & 3.260E-02 & 2.040E-01 \\
$y_{k}$ (m) & 0 & -8.500E-02 & -1.000E-03 & 8.600E-02 & 7.550E-02 & -7.550E-02 & 0 & -3.000E-02 \\
$z_{k}$ (m) & 1.875E-01 & 7.750E-02 & 4.440E-01 & 1.100E-01 & 0 & 8.000E-02 & 0 & 0 \\
\hline
\end{tabular}
\caption{Geometric parameters of the robot links.}
\vspace{-2mm}
\label{geom_param}
\end{table*}

\subsection{Step 5: Identification of Remaining Parameters.}\label{step5}
 
The parameters not identified in the previous operations include the transmission ratio $r_i$, the motor torque constant $K_{t,i}$, the motor back-emf constant $K_{b,i}$, the motor armature resistance $R_i$, the motor armature inductance $L_i$, the rotor inertia $J_{m,i}$, the transmission torsional stiffness $K_{s,i}$, and the transmission damping coefficient $K_{c,i}$.

Although these quantities are necessary for completeness and proper functioning of the model, they are often more difficult to identify than the parameters addressed in the previous subsections. Therefore, they must be estimated based on typical values available in the literature.

Once the parameterization phase is completed, it is necessary to validate the model by comparing and evaluating, from both qualitative and quantitative perspectives, the simulation outputs with real values from experimental measures.

\section{Results of Parameters Identification}\label{param_id_resultd}
\begin{table}[t]
\centering
\begin{tabular}{c|ccc}
\hline
\textbf{$k$-th Link} & $m_k^{CAD}$ (kg) & $\lambda_k$ & $m_k$ (kg) \\
\hline
0 & 4.933 & 1    & 4.933 \\
1 & 3.898 & 1.03 & 4.015 \\
2 & 6.042 & 0.96 & 5.800 \\
3 & 4.767 & 1    & 4.767 \\
4 & 3.915 & 1.03 & 4.032 \\
5 & 4.242 & 1    & 4.242 \\
6 & 0.206 & 1    & 0.206 \\
\hline
\textbf{total} & 28.004 &  & 27.997 \\
\hline
\end{tabular}
\caption{Comparison between CAD masses and identified masses after the correction introduced at Step 2 in Sec. \ref{step2}.}
\label{inertial_param}
\end{table}

The identification procedure of Sec. \ref{identif_sect} is concretely applied here to a real case study. 
The experimental system used in the case study consists of an ABB GoFa CRB15000 collaborative robot, configured to move an extrusion head employed in the 3D printing process, as shown in Fig. \ref{exp_setup}.

The robotic station is equipped with internal sensors that allow the measurement of several fundamental quantities for the identification of the dynamic model. In particular, the robot is able to provide joint positions, motor torques normalized with respect to the transmission ratio, and an estimate of the external torques acting on the joints. These signals are recorded with a maximum sampling frequency of approximately $300$ Hz, thus allowing a relatively detailed analysis of the system dynamics.

For what concerns Step 1 in Sec. \ref{step1},
the following information are available from the CAD file: $x_{g,k},\, y_{g,k},\, z_{g,k}$ (coordinates of the center of mass of the $k$-th link), $x_k,\, y_k,\, z_k$ (coordinates of the next joint), $m_k$ (mass of the $k$-th link), $I_{11,k},\, I_{22,k},\, I_{33,k},\, I_{21,k},\, I_{31,k},\, I_{32,k}$ (independent elements of the symmetric $3 \times 3$ inertia matrix of the $k$-th link), 
$\hat{x}_i,\, \hat{y}_i,\, \hat{z}_i$ (components of the unit vector of the rotation axis of the $i$-th joint). The geometric parameters are reported in Table \ref{geom_param}.

Applying Step 2 in Sec. \ref{step2}, and selecting the correction coefficients $\lambda_k$ in \eqref{lambda_k_eq} so as to minimize the error between estimated and measured torques $\tau_{m,i}r_i$, the estimated inertia parameters and the corresponding corrections are reported in Table \ref{inertial_param}, where it can be observed that the corrections coefficients $\lambda_k$ are all close to one.

\begin{figure}[t]
 \centering\includegraphics[clip,width=\columnwidth]{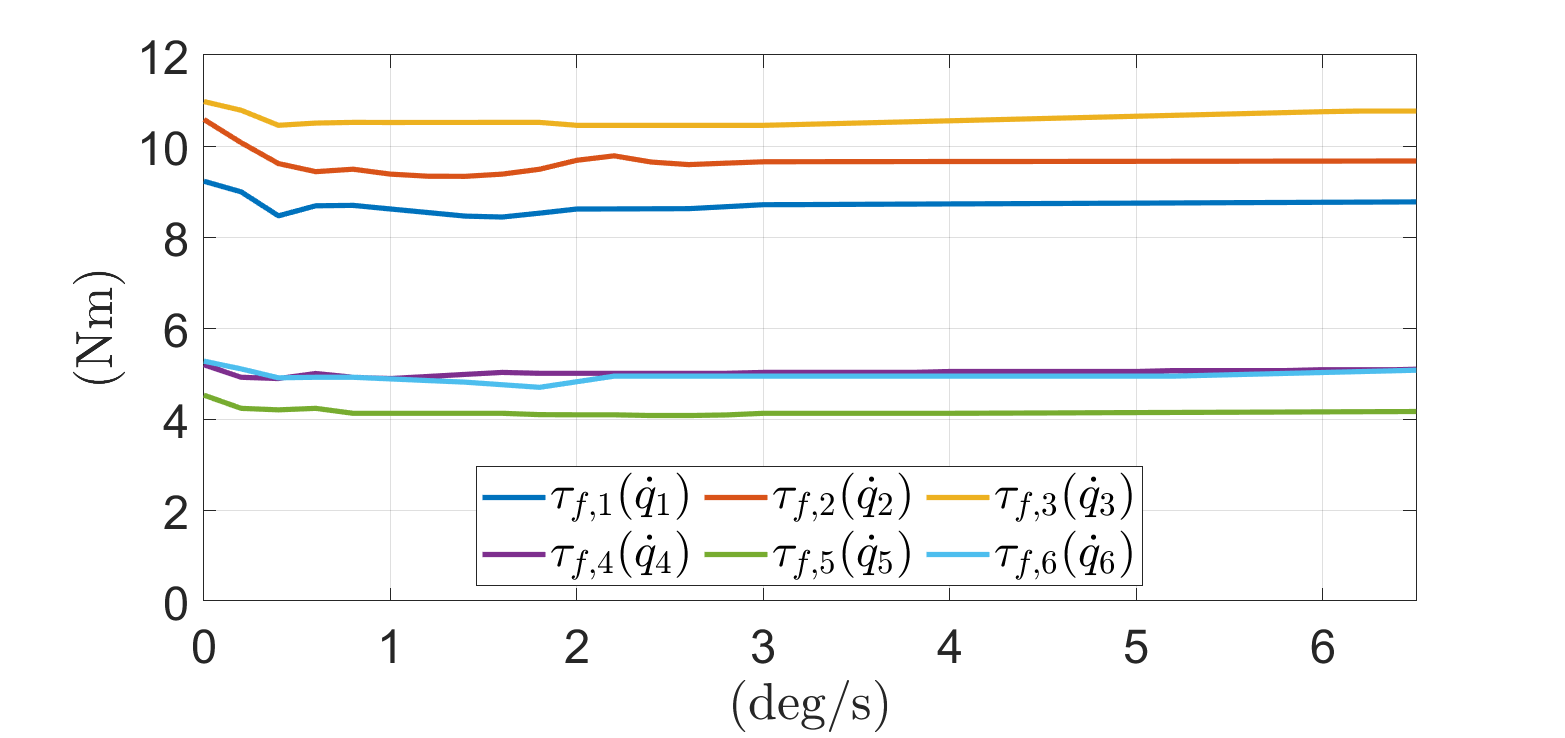}
    \caption{Characteristics of the friction torques $\tau_{f,i}(\dot{q}_i)$.}
\label{friction_torques}
\vspace{-4.2mm} 
 \end{figure}

For what concerns Step 3 in Sec. \ref{step3}, the behavior of the characteristic curves $\tau_{f,i}(\dot{q}_i)$ is shown in Fig. \ref{friction_torques}, while the extrapolated coefficients of the friction model \eqref{sec} are shown in Table \ref{frict_res}.

\begin{table}[t]
\centering
\begin{tabular}{c@{\,}|@{\,}c@{\;}c@{\;}c@{\;}c}
\hline
\textbf{$i$-th Joint} & $F_{s,i}$ (Nm) & $F_{c,i}$ (Nm) & $v'_{s,i}$ (deg/s) & $F_{v,i}$ (Nms/deg) \\
\hline
1 & 9.5  & 8.5  & 0.5 & 0.06 \\
2 & 10.5 & 9.3  & 0.5 & 0.06 \\
3 & 11.5 & 10.5 & 0.5 & 0.06 \\
4 & 5.7  & 4.8  & 0.5 & 0.06 \\
5 & 4.8  & 4.1  & 0.5 & 0.06 \\
6 & 5.4  & 4.8  & 0.5 & 0.06 \\
\hline
\end{tabular}
\caption{Estimated friction parameters for each joint.}
\label{frict_res}
\end{table}

\begin{table}[h!]
\centering
\begin{tabular}{c|ccc}
\hline
\textbf{$i$-th Joint} & $K_{p,i}$ & $T_{z,i}$ & NRMSE (\%) \\
\hline
1 & 1556.2 & 0.2565 & 91.70 \\
2 & 3751.8 & 0.1861 & 92.34 \\
3 & 6823.7 & 0.1489 & 94.84 \\
4 & 3672.5 & 0.1288 & 96.43 \\
5 & 3672.5 & 0.1288 & 94.90 \\
6 & 3672.5 & 0.1288 & 94.11 \\
\hline
\end{tabular}
\caption{Identified controller parameters and corresponding NRMSE values.}
\label{contr_res}
\vspace{-2mm}
\end{table}

For what concerns Step 4 in Sec. \ref{step4}, the 
identified parameters of the controller \eqref{seven} are shown in Table \ref{contr_res}, together with the NRMSE (Normalized Root Mean Squared Error) evaluating the fitting quality.

Finally, for what concerns Step 5 in Sec. \ref{step5}, the choice of $R_i$ and $L_i$ can be reformulated as the choice of $t_{e,i} = L_i/R_i$, the electrical time constant of the motor. The ratio $K_{t,i}/K_{b,i}$ can be assumed constant. It is possible to assume $r_i = 1$ and scale the parameters upstream of the transmission accordingly to this assumption. From a survey of typical values \cite{tang_modeling,chawda2017torque,xie2023antiinertia,bottin2020selectivemodal,xu2021torsionalstiffness}, data for robots similar to the one under investigation are observed.
\begin{table*}[t]
\centering
\scriptsize
\begin{tabular}{c|c@{\;\;\;}c@{\;\;\;}c@{\;\;\;}c@{\;\;\;}c@{\;\;\;}c@{\;\;\;}c@{\;\;\;}c@{\;\;\;}c@{\;\;\;}c@{\;\;\;}c@{\;\;\;}c@{\;\;\;}c@{\;\;\;}c@{\;\;\;}c@{\;\;\;}c@{\;\;\;}c@{\;\;\;}c@{\;\;\;}c}
\hline
\textbf{Point} 
& \cellcolor{green!25}1 & \cellcolor{green!25}2 & 3 & 4 & 5 
& \cellcolor{green!25}6 & \cellcolor{green!25}7 & 8 
& \cellcolor{green!25}9 & \cellcolor{green!25}10 
& 11 & 12 
& \cellcolor{green!25}13 & \cellcolor{green!25}14 
& 15 & 16 & 17 & 18 & 19 \\
\hline

$v_{\mathrm{slow}}$ 
& \cellcolor{green!25}10 & \cellcolor{green!25}20 & 30 & 40 & 100 
& \cellcolor{green!25}10 & \cellcolor{green!25}20 & 30 
& \cellcolor{green!25}10 & \cellcolor{green!25}20 
& 30 & 40 
& \cellcolor{green!25}10 & \cellcolor{green!25}20 
& 30 & 10 & 20 & 40 & 100 \\

$v_{\mathrm{fast}}$ 
& \cellcolor{green!25}100 & \cellcolor{green!25}100 & 100 & 100 & 100 
& \cellcolor{green!25}10 & \cellcolor{green!25}20 & 30 
& \cellcolor{green!25}100 & \cellcolor{green!25}100 
& 100 & 100 
& \cellcolor{green!25}10 & \cellcolor{green!25}20 
& 30 & 100 & 100 & 100 & 100 \\

$a$ 
& \cellcolor{green!25}100\% & \cellcolor{green!25}100\% & 100\% & 100\% & 100\% 
& \cellcolor{green!25}100\% & \cellcolor{green!25}100\% & 100\% 
& \cellcolor{green!25}10\% & \cellcolor{green!25}10\% 
& 10\% & 10\% 
& \cellcolor{green!25}10\% & \cellcolor{green!25}10\% 
& 10\% & 100\% & 100\% & 100\% & 100\% \\

$d$ 
& \cellcolor{green!25}close & \cellcolor{green!25}close & close & close & close 
& \cellcolor{green!25}close & \cellcolor{green!25}close & close 
& \cellcolor{green!25}close & \cellcolor{green!25}close 
& close & close 
& \cellcolor{green!25}close & \cellcolor{green!25}close 
& close & far & far & far & far \\
\hline
\end{tabular}
\caption{Design of Experiments data points.}
\label{points_table}
\end{table*}

\begin{table*}[t]
\centering
\notsotiny
\begin{tabular}{c@{\,}|@{\,}c@{\;\,}c@{\;\,}c@{\;\,}c@{\;\,}c@{\;\,}c@{\;\,}c@{\;\,}c@{\;\,}c@{\;\,}c@{\;\,}c@{\;\,}c@{\;\,}c@{\;\,}c@{\;\,}c@{\;\,}c@{\;\,}c@{\;\,}c@{\;\,}c}
\hline
\textbf{Point}
& 1 & 2 & 3 & 4 & 5 & 6 & 7 & 8 & 9 & 10
& 11 & 12 & 13 & 14 & 15 & 16 & 17 & 18 & 19 \\
\hline
{\tiny \textbf{MAE-X (mm)}}
& 0.0494 & 0.1076 & 0.1588 & 0.2208 & 0.4240
& 0.0433 & 0.0883 & 0.1426 & 0.0461 & 0.0923
& 0.1302 & 0.1528 & 0.0434 & 0.0809 & 0.1237
& 0.0307 & 0.0602 & 0.1160 & 0.2178 \\

{\tiny \textbf{RMSE-X (mm)}}
& 0.1022 & 0.1837 & 0.2477 & 0.3005 & 0.5327
& 0.0798 & 0.1323 & 0.1961 & 0.0860 & 0.1422
& 0.1687 & 0.1998 & 0.0787 & 0.1196 & 0.1650
& 0.0557 & 0.0986 & 0.1538 & 0.2616 \\

{\tiny \textbf{MAE-Y (mm)}}
& 0.0529 & 0.1306 & 0.2067 & 0.2874 & 0.5904
& 0.0463 & 0.1065 & 0.1791 & 0.0479 & 0.1111
& 0.1498 & 0.1881 & 0.0465 & 0.0990 & 0.1544
& 0.1014 & 0.2173 & 0.4443 & 0.8200 \\

{\tiny \textbf{RMSE-Y (mm)}}
& 0.1074 & 0.2204 & 0.3012 & 0.3671 & 0.6849
& 0.0821 & 0.1587 & 0.2343 & 0.0846 & 0.1701
& 0.2128 & 0.2403 & 0.0819 & 0.1451 & 0.1999
& 0.1953 & 0.3629 & 0.5705 & 0.9758 \\
\hline
\end{tabular}
\caption{Error metrics RMSE and MAE between simulation of the identified model and set point.}
\label{simul_metr}
\end{table*}

\begin{table*}[t]
\centering
\notsotiny
\begin{tabular}{c@{\,}|@{\,}c@{\;\,}c@{\;\,}c@{\;\,}c@{\;\,}c@{\;\,}c@{\;\,}c@{\;\,}c@{\;\,}c@{\;\,}c@{\;\,}c@{\;\,}c@{\;\,}c@{\;\,}c@{\;\,}c@{\;\,}c@{\;\,}c@{\;\,}c@{\;\,}c}
\hline
\textbf{Point}
& 1 & 2 & 3 & 4 & 5 & 6 & 7 & 8 & 9 & 10
& 11 & 12 & 13 & 14 & 15 & 16 & 17 & 18 & 19 \\
\hline
{\tiny \textbf{MAE-X (mm)}}
& 0.0511 & 0.1256 & 0.2043 & 0.2921 & 0.4512
& 0.0381 & 0.2346 & 0.2782 & 0.0501 & 0.1654
& 0.2377 & 0.2817 & 0.0437 & 0.2464 & 0.2689
& 0.0412 & 0.4415 & 0.5550 & 0.6149 \\

{\tiny \textbf{RMSE-X (mm)}}
& 0.0792 & 0.2205 & 0.3272 & 0.3907 & 0.5410
& 0.0613 & 0.3061 & 0.3785 & 0.0629 & 0.2158
& 0.3038 & 0.3759 & 0.0617 & 0.3172 & 0.3555
& 0.0587 & 0.5345 & 0.6300 & 0.7579 \\

{\tiny \textbf{MAE-Y (mm)}}
& 0.0312 & 0.1455 & 0.2114 & 0.2909 & 0.3849
& 0.0198 & 0.2711 & 0.3441 & 0.0277 & 0.1776
& 0.2409 & 0.2632 & 0.0210 & 0.2982 & 0.3224
& 0.1053 & 0.3692 & 0.4651 & 0.5243 \\

{\tiny \textbf{RMSE-Y (mm)}}
& 0.0712 & 0.2469 & 0.3403 & 0.3823 & 0.6031
& 0.0360 & 0.3472 & 0.4163 & 0.0444 & 0.2246
& 0.3098 & 0.3615 & 0.0369 & 0.3562 & 0.4015
& 0.1149 & 0.4802 & 0.5651 & 0.7046 \\
\hline
\end{tabular}
\caption{Error metrics RMSE and MAE between robot experimental acquisitions and set point.}
\vspace{-2mm}
\label{exper_metr}
\end{table*}

For $K_{s,i}$, by linearizing around the operating point, the order of magnitude is usually $10^{4}\,\mathrm{Nm/rad}$, with estimates for robots of similar size around $2 \times 10^{4}\,\mathrm{Nm/rad}$, with variations depending on the axis. Typically, the wrist joints (4,5,6) have lower $K_{s,i}$ (by approximately one order of magnitude) \cite{tang_modeling,chawda2017torque,bottin2020selectivemodal,xu2021torsionalstiffness}. Values of $1.5 \times 10^{4}\,\mathrm{Nm/rad}$ are selected for joints 1,2,3 and $1 \times 10^{3}\,\mathrm{Nm/rad}$ for joints 4,5,6.

For $K_{c,i}$, the viscous damping associated with the transmission shows axis-dependent differences similar to those characterizing $K_{s,i}$. Therefore, in agreement with \cite{bottin2020selectivemodal,xu2021torsionalstiffness}, values of $5\,\mathrm{Nms/rad}$ are selected for joints 1,2,3 and $1\,\mathrm{Nms/rad}$ for joints 4,5,6. The resulting transmission damping is often due to nonlinear effects, and it is therefore difficult to assign a precise value to this parameter. Overestimations of $K_{s,i}$ and $K_{c,i}$ ensure better tracking for abrupt trajectories, but in general, for very high values, a further increase does not produce appreciable effects.\\
For $t_{e,i}$, for relatively small motors such as those employed in the system under consideration, a conservative estimate of the electrical time constant is on the order of $10^{-3}\,\mathrm{s}$ 
\cite{chawda2017torque,xie2023antiinertia,xu2021torsionalstiffness}. A value of $1\,\mathrm{ms}$ is selected for all joints.\\
For $J_{m,i}$, from data reported in
\cite{chawda2017torque,bottin2020selectivemodal}, the rotor inertia, expressed upstream of the transmission, is typically on the order of $10^{-3}$ or $10^{-4}\,\mathrm{kg\,m^2}$. Given the assumption $r_i = 1$, this value must be scaled according to typical real transmission ratios (on the order of 50:1 up to approximately 100:1), obtaining values on the order of $1\,\mathrm{kg\,m^2}$. For the wrist joints, in line with the smaller motor size, a value of $0.1\,\mathrm{kg\,m^2}$ is used.

\section{Model Verification through Design of Experiments}\label{Model_Verf}
With reference to Eq. \eqref{first}, it can be observed that: i) for increasing joint velocities $\dot{q}_i$, the term $C_{ij}(q,\dot{q})\,\dot{q}_i$ will become more impactful;  ii) for increasing joint accelerations $\ddot{q}_i$, the term $M_{ij}(q)\,\ddot{q}_i$ will become more impactful; iii) by modifying the position of the printed part with respect to the robot base, the joint positions change, and therefore $g_i(q)$ will change as well.

Therefore, the experiments are designed to vary the following parameters: 
i) the assigned TCP velocity ($v_{\mathrm{slow}}$ for the ``slow'' deposition movement and $v_{\mathrm{fast}}$ for the ``fast'' transition movement between successive depositions); 
ii) the maximum TCP acceleration $a$; 
iii) the distance between the assigned path and the robot base $d$.

\subsection{Parameters Ranges and Tested Conditions}
The lower and upper bounds are selected for the parameters to be varied between one print and another are determined as described in the following. 

For $v_{\mathrm{slow}}$, the lower bound is determined by the printing time, which should not exceed 30–40 minutes. The upper bound is determined by the characteristics of the extruder and typical printing speeds. The parameter is therefore varied from $10\,\mathrm{mm/s}$ to $40\,\mathrm{mm/s}$. 

For $v_{\mathrm{fast}}$, there is no strict lower bound, while the upper bound is determined by robot and controller constraints related to the maximum velocity for short movements. The parameter is therefore varied from $v_{\mathrm{slow}}$ to $100\,\mathrm{mm/s}$. 

For $a$, the lower bound is determined by the controller characteristics, while the upper bound is determined by the robot performance. The employed controller allows modification of this parameter as a percentage of the maximum value. Values below 10\%, for these movements, do not further reduce the acceleration; therefore, $a$ is varied from 10\% to 100\% of the maximum allowable value (which depends on the assigned movements and the robot configuration, but is generally around $4\,\mathrm{m/s^2}$). 

For $d$, there are no actual limits other than those imposed by the robot geometry. Tests are carried out both in the standard printing configuration ($d = d_{\mathrm{close}}$, TCP very close to the base) and in an alternative configuration in which the arm is much more extended ($d = d_{\mathrm{far}}$). The distance between the two positions is $600\,\mathrm{mm}$ on the plane normal to gravity and $300\,\mathrm{mm}$ in height. 

The number of tests is essentially limited by the total printing time. Since the objective is to validate the model rather than investigate the operating point that yields the best quality, a relatively sparse Design of Experiments (DOE) can be formulated to analyze the system behavior at selected key points.

Table \ref{points_table} reports the selected data points for the robot measurements. Out of 19 considered points, 8 (highlighted in green) also involve the actual printing of two cubes per point ($v_{\mathrm{slow}} < 20\,\mathrm{mm/s}$ and $d = d_{\mathrm{close}}$).

\subsection{Validation of the Identified Model}
\begin{figure}[t]
 \centering\includegraphics[clip,width=0.99\columnwidth]{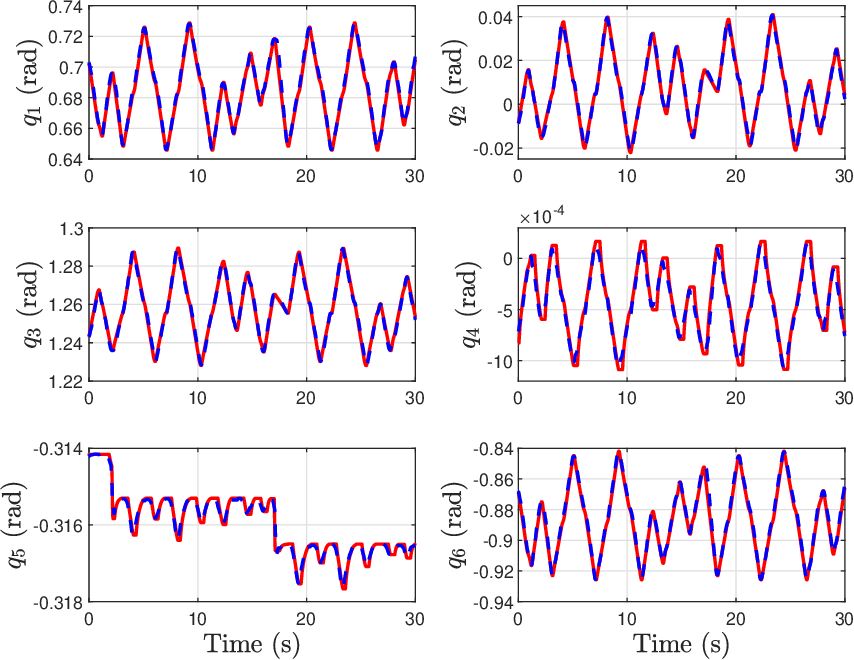}
\vspace{-5.9mm}
    \caption{Joint space trajectories relative to the tenth data point in Table \ref{points_table}: comparison between the identified model (red) and experimental acquisitions (blue).}
\label{Figura_giunti}
 \end{figure}

Fig.~\ref{Figura_giunti} shows the comparison of the joint trajectories from the actual experiments and from the simulation of the identified model. A strong agreement between the experimental and simulation results can be seen, as further confirmed by the Root Mean Squared Error (RMSE) and Mean Absolute Error (MAE) metrics in Table \ref{simul_metr} and Table \ref{exper_metr}, which quantify the error of the TCP coordinates w.r.t the set point.

From Table~\ref{points_table}, Table~\ref{simul_metr} and Table~\ref{exper_metr}, it can be concluded that $v_{\mathrm{slow}}$ is the dominant factor in both simulation and experimental analyses, with higher values leading to a clear degradation of performance, especially along the $y$-direction. The parameter $d$ shows different behaviors: while the model predicts a mixed effect (error increase in $y$ and slight reduction in $x$), experimental data reveal a consistent deterioration in both directions for extended configurations. 
The influence of $a$ and $v_{\mathrm{fast}}$ is limited and statistically weak in both cases.

\begin{table}[t]
\centering
\begin{tabular}{c|ccc}
\hline
\textbf{Point} & max\_out (mm) & max\_in (mm) & diff (mm) \\
\hline
1  & -0.605 & 0.545 & 1.150 \\
2  & -1.310 & 0.781 & 2.091 \\
6  & -0.662 & 0.520 & 1.182 \\
7  & -1.160 & 0.641 & 1.801 \\
9  & -0.566 & 0.421 & 0.987 \\
10 & -1.450 & 0.723 & 2.173 \\
13 & -0.916 & 0.367 & 1.283 \\
14 & -1.290 & 0.591 & 1.881 \\
\hline
\end{tabular}
\caption{Maximum inward and outward deviations and their difference for each of the eight data points of Table \ref{points_table} involving printing.}
\label{cubes_tabl}
\vspace{-4mm}
\end{table}

\begin{figure}[t]
\begin{minipage}{0.48\columnwidth}
    \centering
    \includegraphics[width=0.94\columnwidth]{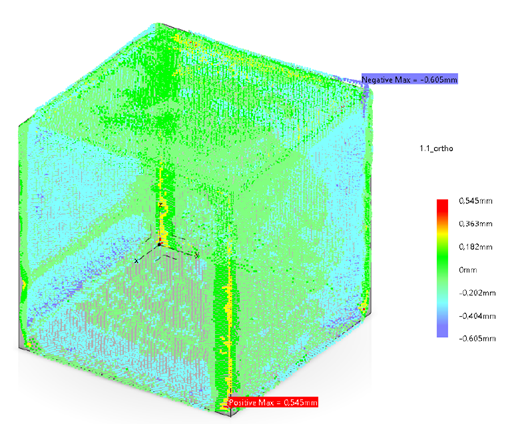}
\end{minipage}
\begin{minipage}{0.48\columnwidth}
    \centering
    \includegraphics[width=0.94\columnwidth]{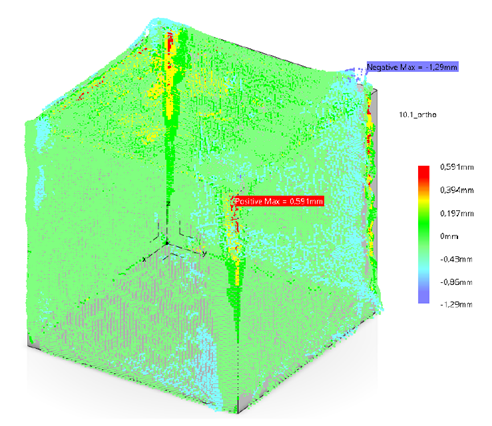}
\end{minipage}
       \put(-193.5,-56){\scriptsize (a)}
       \put(-74,-56){\scriptsize (b)}
       \vspace{-1.25mm}
    \caption{(a) Deviation Analysis for Data Point 1 (lowest errors), and (b) Deviation Analysis for Data Point 14 (most pronounced errors).}
\label{cubes}
\vspace{-1.1mm}
\end{figure}

\subsection{Results on Printed Cubes}
Table \ref{cubes_tabl} shows the maximum absolute deviations (max\_out and max\_in) for each of the eight data points of Table \ref{points_table} involving printing, as well as the sum of the absolute maximum errors in the inward and outward directions of the cube (diff). From Table \ref{points_table} and Table \ref{cubes_tabl}, it can be observed that $v_{\mathrm{slow}}$ has the most significant effect on the quality of the printed product. A clear increase in the error metric is observed as $v_{\mathrm{slow}}$ increases. The effect of $v_{\mathrm{fast}}$ is less straightforward: increasing $v_{\mathrm{fast}}$ improves quality when $v_{\mathrm{slow}} = 20\,\mathrm{mm/s}$, but worsens it when $v_{\mathrm{slow}} = 10\,\mathrm{mm/s}$. This behavior can be partially explained by the fact that, during motion at $v_{\mathrm{fast}}$, the extruder continues processing filament, suggesting that lower values of $v_{\mathrm{fast}}$ should generally lead to poorer print quality. The influence of $a$ is negligible; however, on average, reducing $a$ results in a slight deterioration of print quality. Fig. \ref{cubes} graphically shows the results of the deviation analysis on the points associated with the lowest and highest deviations.

\section{Conclusions}

This work has addressed the proposal of a dynamic modeling approach for  robotic manipulators and of an integrated parameters identification procedure, allowing a reliable dynamic model even in the presence of sensory and programming limitations typical of collaborative robots. The identification procedure involves both the robot dynamics and the actuators and controllers dynamics, and ensures the physical consistency of the identified parameters. The identified model is validated on a real-world case study involving a 6-DoFs collaborative robot used in a thermoplastic extrusion process. The experimental validation confirms a strong agreement between the experimental results and those predicted by the identified model. Future work includes the further refinement of the identified model, to show its impact on enhancing the performance of robot assisted 3D printing.

\section{Acknowledgments}
The authors would like to thank Luca Bosi for his support during this work.

\vspace{2mm}

\bibliographystyle{IEEEtran}
\bibliography{Identification_Col_Robots_3D_Printing}

\end{document}